\documentclass[runningheads]{llncs}
\usepackage[T1]{fontenc}
\usepackage{graphicx,verbatim}
\usepackage{amsmath, amssymb}
\usepackage{booktabs}
\usepackage{multirow}
\usepackage{colortbl}
\usepackage{subcaption}
\usepackage{graphicx}
\begin{document}
\title{PC-MIL: Decoupling Feature Resolution from Supervision Scale in Whole-Slide Learning}

\titlerunning{PC-MIL}
%
\author{%
Syed Fahim Ahmed\inst{1,2}\thanks{Corresponding author: u1419916@utah.edu} \and
Gnanesh Rasineni\inst{1,2} \and
Florian Koehler\inst{3} \and
Abu Zahid Bin Aziz\inst{1,2} \and
Mei Wang\inst{3} \and
Attila Gyulassy\inst{1,2} \and
Brian Summa\inst{4} \and
J. Quincy Brown\inst{4} \and
Valerio Pascucci\inst{1,2} \and
Shireen Y. Elhabian\inst{1,2}
}

\authorrunning{Ahmed et al.}

\institute{%
Kahlert School of Computing, University of Utah, Salt Lake City, UT 84112, USA \and
Scientific Computing \& Imaging Institute, University of Utah, Salt Lake City, UT 84112, USA\\
\email{u1419916@utah.edu, gnanesh.rasineni@utah.edu, zahid.aziz@sci.utah.edu, attila.gyulassy@utah.edu, valerio.pascucci@utah.edu, shireen@sci.utah.edu}
\and
Instapath Inc., Houston, TX 77021, USA\\
\email{florian@instapathbio.com, mwang@instapathbio.com}
\and
Tulane University, Department of Biomedical Engineering, New Orleans, Louisiana, USA\\
\email{bsumma@tulane.edu, jqbrown@tulane.edu}
}
  
\maketitle              
\begin{abstract}
Whole-slide image (WSI) classification in computational pat\-hol\-ogy is commonly formulated as slide-level Multiple Instance Learning (MIL) with a single global bag representation. However, slide-level MIL is fundamentally underconstrained: \textit{optimizing only global labels encourages models to aggregate features without learning anatomically meaningful localization}. This creates a mismatch between the scale of supervision and the scale of clinical reasoning. Clinicians assess tumor burden, focal lesions, and architectural patterns within millimeter-scale regions, whereas standard MIL is trained only to predict whether “somewhere in the slide there is cancer.” As a result, the model's inductive bias effectively erases anatomical structure. 
We propose Progressive-Context MIL (PC-MIL), a framework that treats the spatial extent of supervision as a first-class design dimension. Rather than altering magnification, patch size, or introducing pixel-level segmentation, we decouple feature resolution from supervision scale. Using fixed $20\times$ features, we vary MIL bag extent in millimeter units and anchor supervision at a clinically motivated 2mm scale to preserve comparable tumor burden and avoid confounding scale with lesion density. PC-MIL progressively mixes slide- and region-level supervision in controlled proportions, enabling explicit train-context $\times$ test-context analysis. On 1,476 prostate WSIs from five public datasets for binary cancer detection, we show that anatomical context is an independent axis of generalization in MIL, orthogonal to feature resolution: modest regional supervision improves cross-context performance, and balanced multi-context training stabilizes accuracy across slide and regional evaluation without sacrificing global performance. These results demonstrate that supervision extent shapes MIL inductive bias and support anatomically grounded WSI generalization.

\end{abstract}

\section{Introduction}

Whole-slide image (WSI) analysis is central to computational pathology and has enabled automated cancer assessment at clinical scale \cite{bulten2020automated}. Because WSIs contain billions of pixels, dense pixel-level annotation is rarely available, and weakly supervised learning has become the practical default. Multiple Instance Learning (MIL) formalizes this setting by treating each slide as a bag of patch embeddings supervised only by a global slide label \cite{carbonneau2018multiple,ilse2018attention}.

A key limitation of slide-level MIL is not model capacity but \emph{identifiability}: global labels constrain only \emph{existence} of diagnostic evidence, not its \emph{extent}. Many spatial configurations of evidence are equally consistent with the same slide label, making the learning problem inherently underconstrained. This issue is most apparent in classical MIL aggregators (e.g., max pooling), where gradients can be dominated by a small subset of instances, encouraging shortcut solutions that do not recover anatomically meaningful structure \cite{carbonneau2018multiple}. Attention-based MIL alleviates the rigidity of max pooling by assigning continuous importance to instances \cite{ilse2018attention}, and CLAM further introduces clustering constraints to improve instance-level separability and interpretability \cite{lu2021data}. Nevertheless, these methods still optimize a \emph{global} objective: the only scale at which supervision constrains learning remains the full slide, and spatial extent is not explicitly modeled. When supervision does not constrain spatial extent, the learned representation need not align with anatomically meaningful regions.

This leads to a supervision--reasoning mismatch. Clinical interpretation is inherently spatial—tumor burden, focal lesions, and architectural patterns are assessed within bounded millimeter-scale regions. Indeed, 1–2\,mm thresholds are commonly used to define clinically actionable margins \cite{kuerer2017dcis,scimone2021assessment}. In contrast, slide-level MIL is trained only to determine whether cancer exists \emph{somewhere} in the slide, leaving anatomical support weakly constrained by the learning objective.

A broad literature has improved MIL feature representations and aggregation mechanisms, including transformer-based MIL \cite{shao2021transmil}, dual-stream designs \cite{li2021dual}, and graph-based formulations \cite{li2024dynamic}. In parallel, pathology foundation models have substantially strengthened patch-level embeddings \cite{chen2024towards,vorontsov2024foundation,zimmermann2024virchow2,lu2024visual,KDK2025}, and standardized benchmarks such as EVA enable consistent downstream comparison across encoders \cite{kaiko.ai2024eva}. Multi-scale MIL methods (e.g., SD-MIL) alter perceptual scale or internal feature interactions, yet supervision remains anchored to a single global label. Consequently, the \emph{spatial extent of supervision}—the anatomical scale at which labels constrain the model—has not been treated as an independent design axis shaping MIL inductive bias.

At the same time, many cohorts provide sparse region-of-interest (ROI) annotations that encode localized semantic cues and can be generated efficiently \cite{lu2021data,campanella2019clinical}. While too sparse for dense segmentation training, these ROIs provide a mechanism to impose spatially bounded supervision without requiring pixel-level annotation, enabling weak supervision to be aligned with anatomically meaningful contexts without requiring dense annotations.

We therefore propose Progressive-Context MIL (PC-MIL), which decouples feature resolution from supervision scale by fixing patch representations and explicitly varying the \textit{physical support} over which labels constrain the model. Rather than altering magnification or feature architecture, PC-MIL redefines MIL bags over millimeter-scale anatomical regions while preserving a shared embedding space. Regional supervision is anchored at a clinically motivated 2 mm pivot to maintain comparable tumor burden across contexts and avoid conflating supervision scale with lesion density. By systematically composing slide-level and regional supervision during training and evaluating across contexts, PC-MIL isolates supervision scale as an independent axis of generalization.

We evaluate PC-MIL on 1,476 prostate WSIs aggregated from five public datasets for prostate cancer detection. Our results demonstrate that the supervision scale constitutes an independent axis of generalization in MIL, orthogonal to feature resolution: moderate regional supervision improves cross-context robustness, and balanced multi-context training stabilizes performance across slide-level and regional evaluation scales without sacrificing global accuracy. In summary, we contribute: (i) a principled reframing of MIL in which supervision scale defines inductive bias; (ii) an ROI-guided formulation of millimeter-scale regional supervision anchored at clinically meaningful thresholds; and (iii) a comprehensive empirical study demonstrating anatomically grounded generalization via PC-MIL.

\section{Method}

\begin{figure*}[t]
    \centering
    \includegraphics[width=\linewidth, trim={3mm 2mm 3mm 3mm}, clip]{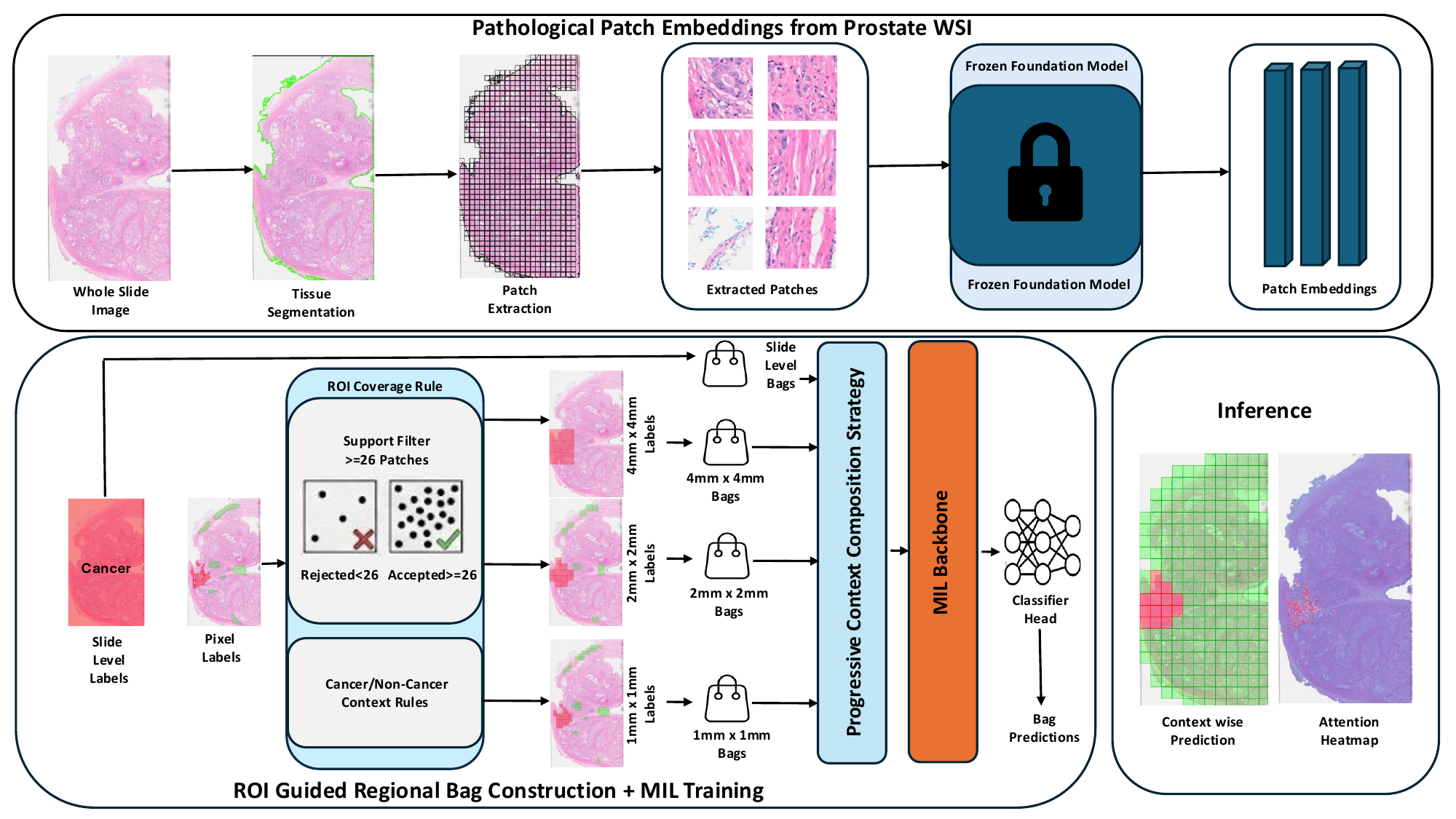}
\caption{
\textbf{PC-MIL pipeline.}
Top: WSIs are segmented, tiled at $20\times$ into $256\times256$ patches, and embedded by a frozen encoder.
Bottom: Sparse ROI annotations generate candidate regional bags across $4\times4, 2\times2,$ and $1\times1\,\text{mm}^2$ anatomical extents using coverage rules (red: Cancer, green: Non-Cancer; ambiguous regions are discarded).
Each WSI is assigned to one supervision context during training to prevent cross-context leakage, and contexts are mixed via $\boldsymbol{\alpha}$.
Inference produces context-wise predictions and attention heatmaps.
}
    \label{fig:pipeline}
\end{figure*}

We introduce Progressive-Context MIL (PC-MIL) to study supervision scale as a controllable design variable in MIL while preserving a shared embedding space. PC-MIL explicitly varies the anatomical extent of MIL bags without altering feature resolution. The framework comprises three components: (1) ROI-guided regional bag construction anchored at a clinically motivated $2,\text{mm}$ pivot; (2) a granularity-first context allocation strategy to prevent cross-context information leakage; and (3) a progressive context-mixing scheme governed by a composition vector $\boldsymbol{\alpha}$ that enables learning across multiple anatomical scales. An overview of the full PC-MIL pipeline is shown in Fig.~\ref{fig:pipeline}.

\vspace{0.05in}
\noindent\textbf{Background: Slide-Level Multiple Instance Learning}. In computational pathology, WSIs are partitioned into patches and analyzed using Multiple Instance Learning (MIL). We define a dataset of $N$ WSIs as $\mathcal{S}=\{(S_i,Y_i)\}_{i=1}^{N}$, where each slide $S_i$ has a binary label $Y_i\in\{0,1\}$. In MIL \cite{ilse2018attention}, each slide is a bag $B_i=\{x_{ij}\}_{j=1}^{M_i}$ of $M_i$ patches. Under the classical binary MIL assumption, the bag label is positive iff at least one instance is positive, $Y_i = \max_{j} y_{ij}$, where instance labels $y_{ij}\in\{0,1\}$ are \emph{latent}. Each patch is embedded with a frozen foundation encoder to generate features $\mathbf{f}_{ij}=\phi(x_{ij})\in\mathbb{R}^{D}$. To explicitly define that aggregation produces a global embedding rather than a direct label, we utilize a permutation-invariant function $g(\cdot)$ and a classifier head $h(\cdot)$:
\begin{equation}
\mathbf{z}_i = g\big(\{\mathbf{f}_{ij}\}_{j=1}^{M_i}\big), \quad \hat{p}_i = \sigma(h(\mathbf{z}_i)), \quad \hat{Y}_i = \mathbb{I}\left[\hat{p}_i \ge \tau\right],
\end{equation}
where $\mathbf{z}_i$ is the \emph{slide-level embedding}, $\sigma(\cdot)$ is the sigmoid activation function, and $\hat{p}_i$ is the predicted probability. In conventional MIL, the entire WSI constitutes a single bag, $B_i^{\text{slide}}=\{\mathbf{f}_{ij}\}_{j=1}^{M_i}$, so aggregation is performed globally. In this formulation, supervision constrains only the \emph{existence} of positive evidence, not the anatomical extent. In this work, we generalize the bag definition to anatomically defined regional extents, enabling explicit control over the spatial extent of supervision.

\vspace{0.05in}
\noindent\textbf{ROI-Guided Regional Bag Construction}. Fig.~\ref{fig:pipeline} summarizes the ROI-guided multi-scale bag construction and context allocation strategy. Conventional MIL treats an entire WSI as a single bag, implicitly aggregating information across the full tissue extent. In contrast, we redefine the spatial extent of a bag by processing WSIs at fixed $20\times$ magnification ($\approx 0.5\,\mu\text{m/pixel}$) using non-overlapping $256 \times 256$ patches. This fixed tiling ensures that millimeter-scale regions correspond to disjoint, countable patch sets, enabling consistent geometric partitioning of the slide and precise control over absolute tumor burden across supervision scales. At $20\times$ resolution, square regional contexts of $1\times1$, $2\times2$, and $4\times4\,\text{mm}^2$ correspond to pixel extents of $2048^2$, $4096^2$, and $8192^2$, respectively, which translate to bags of $64$, $256$, and $1024$ patches.

Each slide $S_i$ is partitioned into non-overlapping square regions of size $c \in \{1\times1, 2\times2, 4\times4\}\,\text{mm}^2$.
Let $n_{i,r}^{\text{tumor}}, n_{i,r}^{\text{normal}}, n_{i,r}^{\text{stroma}}$ denote the number of annotated patches
of each type in region $r$, and $n_{i,r}^{\text{ann}}$ their sum.
A region is ROI-valid if $n_{i,r}^{\text{ann}}\ge 26$. A valid region is labeled \textit{Cancer} if
$n_{i,r}^{\text{tumor}}\ge 26$; it is labeled \textit{Non-Cancer} if $n_{i,r}^{\text{tumor}}=0$ and
either $n_{i,r}^{\text{normal}}/n_{i,r}^{\text{ann}}\ge 0.5$ or $n_{i,r}^{\text{stroma}}=n_{i,r}^{\text{ann}}$.
Valid regions that do not meet either criterion are treated as ambiguous and discarded.
The threshold of 26 patches corresponds to a fixed absolute tumor area under the $2$mm pivot and is inspired by
millimeter-scale margin criteria in surgical pathology \cite{kuerer2017dcis,scimone2021assessment}.

We define the set of supervision contexts as $\mathcal{C}=\{\text{slide},\,1\times1,\,2\times2,\,4\times4\}\ \text{mm}^2$, where $c\in\{1\times1,\,2\times2,\,4\times4\}$ denotes square regions of side length $1, 2,$ or $4$\,mm. For any $c \in \mathcal{C}$, the collection of bags for slide $S_i$ is $\mathcal{B}_i^{(c)} = \{B_i^{\text{slide}}\}$ if $c=\text{slide}$, and $\mathcal{B}_i^{(c)} = \{B_{i,r}^{(c)} \mid r \text{ is ROI-valid at scale } c\}$ for regional contexts. A regional bag aggregates patch embeddings within a valid region $r$, defined as $B_{i,r}^{(c)}=\{\mathbf{f}_{ij}\mid x_{ij}\in r\}$. Unlike the slide bag which includes all $M_i$ patches, regional bags are restricted to ROI-validated contexts to ensure high-confidence supervision derived from sparse annotations. We denote the corresponding bag label as $Y_{i,r}^{(c)}$, where $Y_i$ is the ground truth for $c=\text{slide}$.

\vspace{0.05in}
\noindent\textbf{Progressive-Context MIL (PC-MIL).} We treat the supervision scale as a controllable design variable by mixing labels across contexts $\mathcal{C}=\{\text{slide}, 4\times4, 2\times2, 1\times1\}\,\text{mm}^2$. We define a \emph{supervision composition vector} $\boldsymbol{\alpha}=(\alpha_{\text{slide}}, \alpha_{4\times4}, \alpha_{2\times2}, \alpha_{1\times1})$ satisfying $\sum_{c\in\mathcal{C}}\alpha_c=1$, which specifies the fraction of WSIs assigned to each anatomical scale.
At dataset construction, each slide $S_i$ is assigned a single context $c_i$ according to $\boldsymbol{\alpha}$,
and training uses only bags $\mathcal{B}_i^{(c_i)}$ from that scale.
This granularity-first allocation ensures that a WSI contributes supervision at only one anatomical scale,
preventing cross-context leakage.

After context allocation, we balance Cancer and Non-Cancer bags.
For each bag $B\in\mathcal{B}_i^{(c_i)}$, the permutation-invariant aggregator produces a context embedding
$\mathbf{z}=g(B)$, and a classification head predicts $\hat{p}=\sigma(h(\mathbf{z}))$,
which is supervised using the corresponding bag label.

\vspace{0.05in}
\noindent\textbf{Experimental Design and Context Sweep.}
\label{subsec:Context-Composition}
To quantify supervision-scale effects, we sweep $\boldsymbol{\alpha}$ by reducing the slide fraction from $100\%$ to $60\%$ while reallocating mass to the $4\times4$, $2\times2$, and $1\times1\,\text{mm}^2$ square regional contexts. We evaluate each trained model under fixed inference contexts $c^\ast\in\mathcal{C}$ to perform a controlled train-context $\times$ test-context analysis (Sec.~\ref{sec:Results}).

\vspace{0.05in}
\noindent\textbf{Implementation Details.}
We use five public prostate H\&E datasets: AGGC22 \cite{huo2024comprehensive}, TCGA-PRAD \cite{zuley2016tcgaprad},
GTEx-Prostate \cite{centergtex}, DiagSet \cite{koziarski2024diagset}, and Gleason2019 \cite{nir2018automatic}.
Official train/val/test splits are used for AGGC22, DiagSet, and Gleason2019; TCGA-PRAD and GTEx are split 80/10/10 at
the slide level. All regional bags inherit the split of their parent slide.

WSIs are processed using TRIDENT \cite{zhang2025accelerating} 
(tissue masking; $20\times$ patching at $256\times256$). 
Patches are embedded using a frozen UNI2-h encoder into 
$\mathbf{f}_{ij}\in\mathbb{R}^{1536}$; the encoder remains fixed across experiments 
so performance differences arise solely from bag construction and supervision scale.

MIL backbones are implemented using the MIL-LAB framework \cite{shao2025multiple}, 
which provides standardized implementations of multiple state-of-the-art 
architectures. We evaluate ABMIL \cite{ilse2018attention}, TransMIL 
\cite{shao2021transmil}, DSMIL \cite{li2021dual}, WiKG \cite{li2024dynamic}, 
RRT \cite{tang2024feature}, Transformer-based MIL \cite{ilse2018attention}, 
and ILRA \cite{xiang2023exploring} under identical optimization settings.

We specifically evaluate two ABMIL variants: (1) \textbf{ABMIL (pre)}, representing the standard ABMIL \cite{ilse2018attention} initialized with weights from \cite{shao2025multiple}; and (2) \textbf{PC-MIL}, which implements our progressive context-mixing strategy using a vanilla ABMIL backbone using random weights (denoting as \textbf{PC-MIL}). Let $\mathcal{T}$ denote the set of training bags after context allocation 
and balancing, with $\hat{p}_B=\sigma(h(g(B)))$. 
We optimize bag-level binary cross-entropy:
\begin{equation}
\mathcal{L}_{\text{BCE}}
=
-\frac{1}{|\mathcal{T}|}
\sum_{(B, Y_B) \in \mathcal{T}}
\left[
Y_B \log \hat{p}_B
+
(1 - Y_B)\log (1 - \hat{p}_B)
\right].
\end{equation}
Optimization uses AdamW with early stopping on validation balanced accuracy.
Evaluation is performed separately under each fixed inference context 
$c^\ast\in\mathcal{C}$.

\section{Results}
\label{sec:Results}

In prostate cancer diagnostics, high specificity at 98\% sensitivity (S@98) is clinically critical to reduce unnecessary biopsies while minimizing missed clinically significant disease \cite{Mazzone2022,Trevino2021}. We therefore report balanced accuracy (B-A), which accounts for class imbalance, and S@98 across both slide-level and regional inference contexts to evaluate cross-scale generalization.

\begin{table}[t]
\centering
\caption{Binary classification performance across progressive multi-context training label allocations. Training ratios $\boldsymbol{\alpha}$: (Slide, $4\times4, 2\times2, 1\times1$)\,\%. B-A: Balanced Accuracy (\%); S@98: Spec. at 98\% Sens. (\%). Average columns denote mean performance across all four evaluation contexts $c^\ast \in \mathcal{C}$.}
\label{tab:multi_context_final}
\scriptsize
\setlength{\tabcolsep}{0pt}
\begin{tabular*}{\textwidth}{@{\extracolsep{\fill}} l cc cc cc cc cc @{}}
\toprule
\textbf{Ratio $\boldsymbol{\alpha}$} & \multicolumn{2}{c}{\textbf{Slide}} & \multicolumn{2}{c}{\textbf{$4\times4\,\text{mm}^2$}} & \multicolumn{2}{c}{\textbf{$2\times2\,\text{mm}^2$}} & \multicolumn{2}{c}{\textbf{$1\times1\,\text{mm}^2$}} & \multicolumn{2}{c}{\textbf{Average}} \\
\cmidrule(lr){2-3} \cmidrule(lr){4-5} \cmidrule(lr){6-7} \cmidrule(lr){8-9} \cmidrule(l){10-11}
(\%) & \textbf{B-A} & \textbf{S@98} & \textbf{B-A} & \textbf{S@98} & \textbf{B-A} & \textbf{S@98} & \textbf{B-A} & \textbf{S@98} & \textbf{B-A} & \textbf{S@98} \\
\midrule
(100,0,0,0)  & \textbf{99.46} & \textbf{99.73} & 78.78 & 85.49 & 69.00 & 93.98 & 63.09 & 89.42 & 77.58 & 92.16 \\
(90,10,0,0)  & \textbf{99.46} & \underline{99.46} & \underline{96.15} & 94.08 & 93.56 & 89.62 & 87.14 & 77.74 & 94.08 & 90.22 \\
(90,6,4,0)   & 98.66 & 97.85 & 95.49 & 95.27 & 91.31 & 95.48 & 86.26 & 89.64 & 92.93 & 94.56 \\
(90,6,2,2)   & 98.39 & 97.58 & 94.96 & 89.55 & \textbf{95.31} & 95.68 & \underline{95.71} & 98.04 & \underline{96.09} & 95.21 \\
(80,20,0,0)  & 98.12 & 97.58 & 90.36 & 90.37 & 88.33 & 92.38 & 84.44 & 90.12 & 90.31 & 92.61 \\
(80,12,8,0)  & 98.39 & \underline{99.46} & 95.98 & 86.23 & 93.65 & 96.09 & 90.33 & 96.93 & 94.59 & 94.68 \\
(80,12,4,4)  & 98.12 & 95.13 & 95.65 & \textbf{96.47} & 94.26 & \textbf{98.10} & 92.26 & 97.84 & 95.07 & \textbf{96.89} \\
(70,30,0,0)  & 98.12 & 99.18 & 88.47 & 81.81 & 84.18 & 93.02 & 79.21 & 92.11 & 87.50 & 91.53 \\
(70,18,12,0) & 97.83 & 95.18 & \textbf{97.05} & \underline{96.31} & \underline{95.16} & \underline{97.54} & 91.93 & 97.21 & 95.49 & \underline{96.56} \\
(70,18,6,6)  & 98.12 & 97.85 & 88.86 & 80.62 & 91.07 & 96.72 & 94.26 & \underline{98.16} & 93.08 & 93.34 \\
(60,40,0,0)  & 96.49 & 96.49 & 93.02 & 89.32 & 88.04 & 95.51 & 82.20 & 94.32 & 89.94 & 93.91 \\
(60,24,16,0) & 97.33 & 95.70 & 94.31 & 87.93 & 92.82 & 96.96 & 91.92 & 97.29 & 94.09 & 94.47 \\
(60,24,8,8)  & 97.31 & 96.76 & 95.05 & 85.20 & 95.01 & 94.91 & \textbf{97.26} & \textbf{98.18} & \textbf{96.16} & 93.76 \\
\bottomrule
\end{tabular*}
\smallskip
{\scriptsize\raggedright\textbf{Bold} indicates best performance per evaluation context; \underline{underlined} indicates second best.\par}
\end{table}

Table~\ref{tab:multi_context_final} reveals a strong supervision-scale dependence in cross-context performance. 
A slide-only model ($100,0,0,0$) attains near-ceiling slide accuracy ($99.46\%$ B-A) but exhibits severe degradation under regional inference ($69.00\%$ and $63.09\%$ B-A at $2\times2$ and $1\times1\,\text{mm}^2$ square regions, respectively), consistent with the underconstrained nature of slide-level MIL.Injecting limited regional supervision yields substantial cross-context gains: allocating only $10\%$ of slides to $4\times4\,\text{mm}^2$ square regions ($90,10,0,0$) increases average B-A from $77.58\%$ to $94.08\%$ while preserving slide performance. Incorporating finer contexts further stabilizes performance across scales; $(60,24,8,8)$ achieves the best average B-A ($96.16\%$), while $(80,12,4,4)$ yields the best average S@98 ($96.89\%$).These results suggest a stable regime in which moderate regional supervision maximizes cross-context robustness without materially compromising global accuracy. Beyond this regime, increasing regional supervision produces diminishing returns and may slightly reduce slide-level performance.

\begin{table}[t]
\centering
\caption{Comparison of PC-MIL against state-of-the-art baselines across multi-context evaluation. B-A: Balanced Accuracy (\%); S@98: Spec. at 98\% Sens. (\%). Average columns denote mean performance across all four evaluation contexts.}
\label{tab:baseline_comparison}
\scriptsize
\setlength{\tabcolsep}{0pt}
\begin{tabular*}{\textwidth}{@{\extracolsep{\fill}} l cc cc cc cc cc @{}}
\toprule
\textbf{Model} & \multicolumn{2}{c}{\textbf{Slide}} & \multicolumn{2}{c}{\textbf{$4\times4\,\text{mm}^2$}} & \multicolumn{2}{c}{\textbf{$2\times2\,\text{mm}^2$}} & \multicolumn{2}{c}{\textbf{$1\times1\,\text{mm}^2$}} & \multicolumn{2}{c}{\textbf{Average}} \\
\cmidrule(lr){2-3} \cmidrule(lr){4-5} \cmidrule(lr){6-7} \cmidrule(lr){8-9} \cmidrule(l){10-11}
& \textbf{B-A} & \textbf{S@98} & \textbf{B-A} & \textbf{S@98} & \textbf{B-A} & \textbf{S@98} & \textbf{B-A} & \textbf{S@98} & \textbf{B-A} & \textbf{S@98} \\
\midrule
ABMIL (pre) \cite{shao2025multiple}  & 98.91 & 97.85 & 77.66 & 67.62 & 66.49 & 64.89 & 60.66 & 63.71 & 75.93 & 73.52 \\
TransMIL~\cite{shao2021transmil}     & 98.66 & \underline{99.46} & 79.36 & 89.89 & 77.80 & 92.83 & 80.99 & 92.04 & 84.20 & 93.56 \\
DSMIL~\cite{li2021dual}        & 98.64 & 99.18 & 78.51 & 66.24 & 69.85 & 85.03 & 64.78 & 83.86 & 77.94 & 83.58 \\
WiKG~\cite{li2024dynamic}         & 97.85 & 97.58 & 77.66 & 66.05 & 66.81 & 72.69 & 61.14 & 72.32 & 75.87 & 77.16 \\
RRT~\cite{tang2024feature}          & \underline{99.18} & 99.18 & 79.30 & 87.11 & 68.85 & 88.15 & 70.58 & 84.19 & 79.48 & 89.66 \\
Transformer~\cite{ilse2018attention}  & 97.85 & 97.85 & 78.26 & 88.77 & 68.75 & 85.02 & 62.70 & 75.83 & 76.89 & 86.87 \\
ILRA~\cite{xiang2023exploring}         & 98.12 & 99.18 & 77.71 & 31.21 & 65.93 & 23.65 & 59.67 & 22.02 & 75.36 & 44.02 \\
ABMIL~\cite{ilse2018attention}        & \textbf{99.46} & \textbf{99.73} & 78.78 & 85.49 & 69.00 & 93.98 & 63.09 & 89.42 & 77.58 & 92.16 \\
\midrule
PC-MIL (Slide-Pres) & \textbf{99.46} & \underline{99.46} & \textbf{96.15} & \underline{94.08} & 93.56 & 89.62 & 87.14 & 77.74 & 94.08 & 90.22 \\
PC-MIL (Clinical)   & 98.12 & 95.13 & \underline{95.65} & \textbf{96.47} & \underline{94.26} & \textbf{98.10} & \underline{92.26} & \underline{97.84} & \underline{95.07} & \textbf{96.89} \\
PC-MIL (Balanced)   & 97.31 & 96.76 & 95.05 & 85.20 & \textbf{95.01} & \underline{94.91} & \textbf{97.26} & \textbf{98.18} & \textbf{96.16} & \underline{93.76} \\
\bottomrule
\end{tabular*}
\smallskip
{\scriptsize\raggedright\textbf{Bold} indicates best performance per evaluation context; \underline{underlined} indicates second best.\par}
\end{table}

Based on Table~\ref{tab:multi_context_final}, we select three representative allocations: \textit{Slide-Preserving} ($90,10,0,0$), \textit{Clinical} ($80,12,4,4$; best avg. S@98), and \textit{Balanced} ($60,24,8,8$; best avg. B-A). Table~\ref{tab:baseline_comparison} shows that architectural variation alone does not prevent cross-context degradation: slide-level baselines remain strong on slide evaluation but drop markedly on regional square contexts (avg. B-A 75.36--84.20\%). In contrast, PC-MIL improves robustness across scales. \textit{Slide-Preserving} retains 99.46\% slide B-A while raising $4\times4\,\text{mm}^2$ B-A to 96.15\% (94.08\% avg. B-A). \textit{Balanced} yields the best avg. B-A (96.16\%), while \textit{Clinical} achieves the best average S@98 (96.89\%). Overall, these results indicate that the supervision scale is an independent driver of inductive bias and cross-context generalization in MIL, with a larger impact than backbone choice in this setting.

\begin{figure}[t]
    \centering
    \begin{subfigure}{\linewidth}
        \centering
        \label{fig:2mm_preds}
        \includegraphics[width=\linewidth, trim={1cm 8cm 1cm 4cm}, clip]{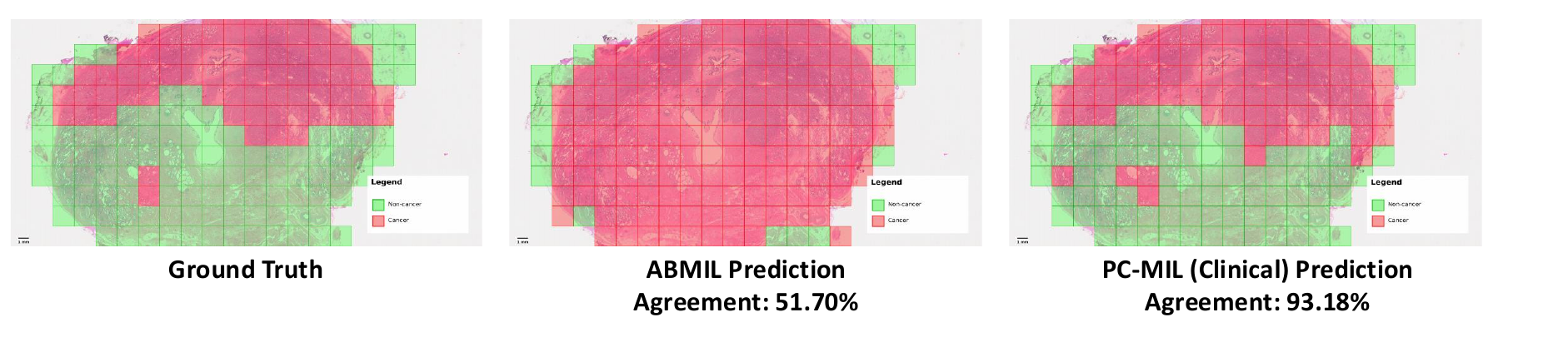}
        \caption{$2\times2\,\text{mm}^2$ regional predictions comparing ABMIL and PC-MIL.}
        \label{fig:2mm_preds}
    \end{subfigure}
    \begin{subfigure}{\linewidth}
        \centering
        \label{fig:attention_heatmaps}
        \includegraphics[width=\linewidth, trim={1cm 9cm 1cm 4cm}, clip]{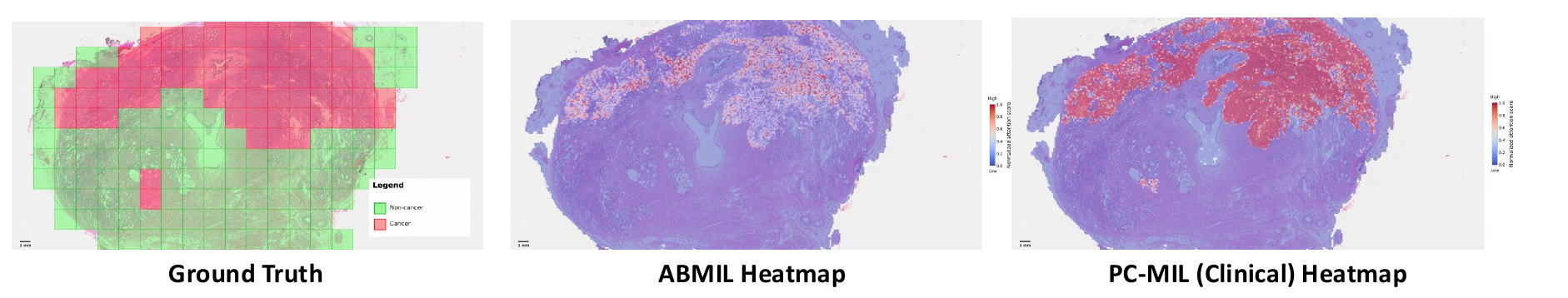}
        \caption{Corresponding attention heatmaps.}
        \label{fig:attention_heatmaps}
    \end{subfigure}
    
    \caption{Qualitative comparison of spatial reasoning.
    }
    \label{fig:2mm_vs_heatmap}
    \vspace{-1.5em}
\end{figure}

Figure~\ref{fig:2mm_vs_heatmap} provides qualitative support for the mechanism suggested by Tables~\ref{tab:multi_context_final}--\ref{tab:baseline_comparison}. When trained solely with slide-level supervision, ABMIL produces overextended 2mm cancer predictions and broadly distributed attention, consistent with the underconstrained nature of slide-level MIL: correct slide classification can be achieved without learning anatomically coherent tumor extent. In contrast, PC-MIL trained with 2mm regional supervision yields spatially consistent predictions that better align with the ground-truth tumor distribution (93.18\% vs 51.70\% agreement), concentrating high-confidence responses within tumor-dense regions rather than diffusely across peripheral tissue.
To establish regional ground truth for this evaluation, 10 WSIs from the AGGC22 dataset \cite{huo2024comprehensive} were exhaustively annotated at $2\times2\,\text{mm}^2$ square regions by expert pathologists using a custom regional annotation tool. Regional inference was performed across the entire tissue, and agreement was computed against these localized annotations. For visualization only, $2\times2\,\text{mm}^2$ region probabilities were binarized at a conservative threshold of 0.9 to highlight high-confidence cancer regions; quantitative evaluation used continuous predictions.
These qualitative patterns mirror the quantitative improvements observed in cross-context evaluation and provide further evidence that supervision scale shapes the inductive bias of MIL, promoting anatomically grounded reasoning beyond what is achieved through architectural modification alone.

\section{Conclusion}
\label{sec:conclusion}

We presented Progressive-Context MIL (PC-MIL), which treats the spatial extent of supervision as a first-class design variable for WSI classification. Using fixed $20\times$ features, PC-MIL decouples feature resolution from supervision scale and progressively mixes slide-level and millimeter-scale regional supervision to analyze train-context $\times$ test-context interactions. Across 1,476 prostate WSIs from five public datasets, we show that supervision scale is an independent axis of generalization: slide-only training attains near-ceiling slide accuracy but degrades under regional inference, while modest regional supervision yields substantial cross-context gains and balanced multi-context training stabilizes performance across slide and regional evaluations without sacrificing global accuracy. Qualitative predictions and attention maps further indicate that regional supervision encourages anatomically coherent localization rather than diffuse aggregation. Future work will extend PC-MIL to multi-class Gleason grading and study calibration and adaptive context allocation for deployment.

\bibliographystyle{splncs04}
\bibliography{reference}

\end{document}